\documentclass[letterpaper, 10 pt, conference]{ieeeconf}  % Comment this line out if you need a4paper

\IEEEoverridecommandlockouts                              % This command is only needed if 
\usepackage{subfigure}
\usepackage[utf8]{inputenc} % allow utf-8 input
\usepackage[T1]{fontenc}    % use 8-bit T1 fonts
\usepackage{hyperref}       % hyperlinks
\usepackage{url}            % simple URL typesetting
\usepackage{booktabs}       % professional-quality tables
\usepackage{amsfonts}       % blackboard math symbols
\usepackage{nicefrac}       % compact symbols for 1/2, etc.
\usepackage{microtype}      % microtypography
\usepackage{xcolor}         % colors
\usepackage{graphicx}
\usepackage{multirow}
\usepackage{tabularx}
\usepackage{xspace}
\usepackage{cite}

\newcommand{\tio}{TiO\xspace}

\title{\LARGE \bf
Towards Unified Interactive Visual Grounding in The Wild
}

\author{Jie Xu$^{\dag1,2}$, Hanbo Zhang*$^2$, Qingyi Si$^{2}$, Yifeng Li$^2$, Xuguang Lan*$^1$, and Tao Kong$^2$
\thanks{$^1$ Xi'an Jiaotong University, $^2$ ByteDance Research}
\thanks{$^\dag$ This work was done during an internship at ByteDance.}
\thanks{* Correspondence to: Hanbo Zhang \{zhb@bytedance.com\} and Xuguang Lan \{xglan@mail.xjtu.edu.cn\}}
}

\begin{document}

\maketitle
\thispagestyle{empty}
\pagestyle{empty}

%%%%%%%%%%%%%%%%%%%%%%%%%%%%%%%%%%%%%%%%%%%%%%%%%%%%%%%%%%%%%%%%%%%%%%%%%%%%%%%%

\begin{abstract}
Interactive visual grounding in Human-Robot Interaction (HRI) is challenging yet practical due to the inevitable ambiguity in natural languages.
It requires robots to disambiguate the user's input by active information gathering.
% Existing solutions are to apply three independent agents, i.e., Questioner, Oracle and Guesser, which separate visual dialog and grounding completely.
Previous approaches often rely on predefined templates to ask disambiguation questions, resulting in performance reduction in realistic interactive scenarios.
In this paper, we propose \tio, an end-to-end system for interactive visual grounding in human-robot interaction.
% inspired by the human experience that the best way to avoid ambiguity is to hold a dialog and ask questions in person rather than listening to others. 
% To this end, we formulate a novel paradigm that applies a single model to play all three agents, and accordingly propose a unified visual dialog and grounding transformer.
Benefiting from a unified formulation of visual dialog and grounding, our method can be trained on a joint of extensive public data, and show superior generality to diversified and challenging open-world scenarios.
In the experiments, we validate \tio on GuessWhat?! and InViG benchmarks, setting new state-of-the-art performance by a clear margin. 
% Besides, we conduct a comprehensive analysis to show why \\tio can significantly improve the performance of interactive visual-language disambiguation.
Moreover, we conduct HRI experiments on the carefully selected 150 challenging scenes as well as real-robot platforms. 
Results show that our method demonstrates superior generality to diversified visual and language inputs with a high success rate.
Codes and demos are available on \href{https://github.com/jxu124/TiO}{https://github.com/jxu124/TiO}.
  
  % a Extensive experiments validate the superiority of the proposed paradigm which outperforms the state-of-the-art on both InViG and GuessWhat?! datasets, and d 
% We also conduct comprehensive ablations of each , and systematically study the roles of varying types of knowledge. 
\end{abstract}

%%%%%%%%%%%%%%%%%%%%%%%%%%%%%%%%%%%%%%%%%%%%%%%%%%%%%%%%%%%%%%%%%%%%%%%%%%%%%%%%

\section{Introduction}

\begin{figure}
  \centering
  % \fbox{\rule[-.5cm]{0cm}{4cm} \rule[-.5cm]{4cm}{0cm}}
  \includegraphics[width=1\columnwidth]{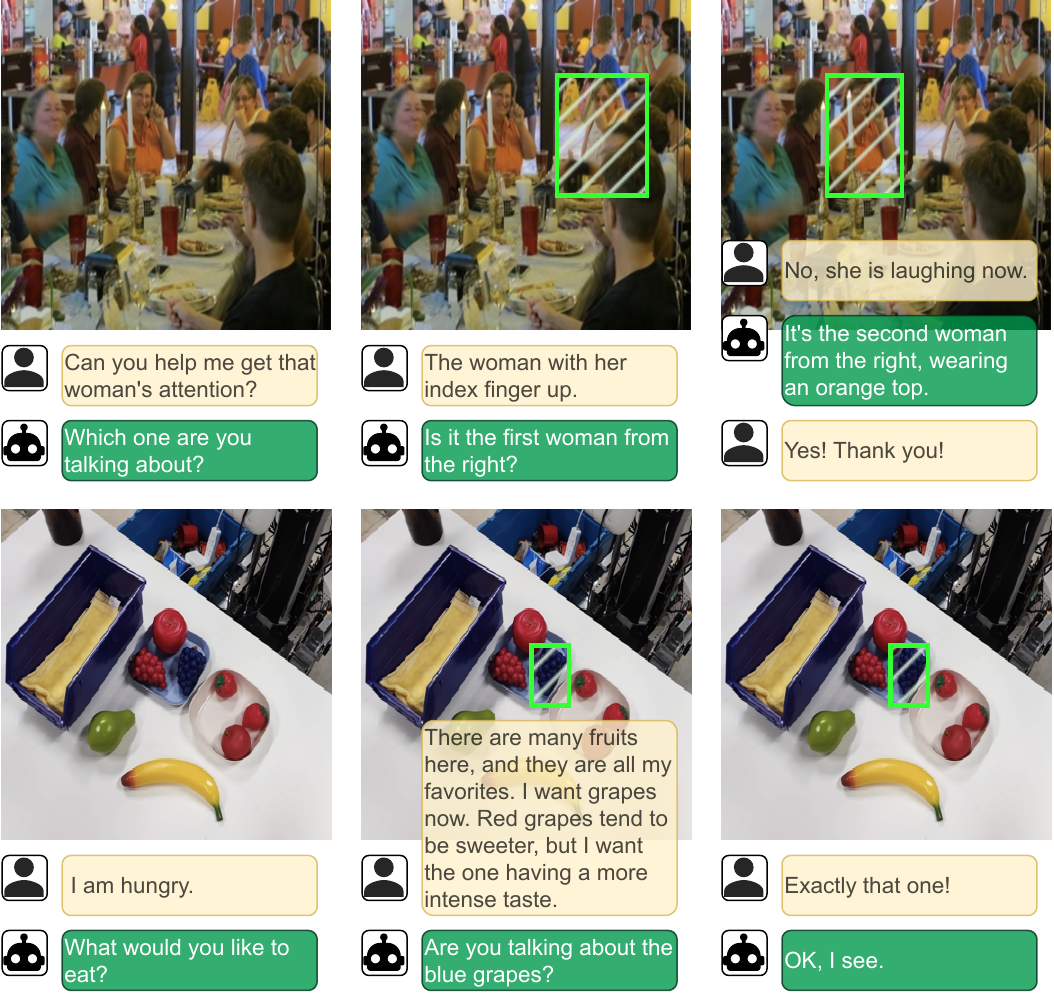} 
  
  \caption{\tio in the wild (top row) and the realistic interactive robot manipulation tasks (bottom row). Each image along with the corresponding round of in the dialog shows that \tio can ask informative questions based on previous dialog history and complex observations, while maintaining explainable internal states to evaluate the grounded candidates (green box) of the target.}
    \label{fig:fig_1}
\end{figure}

% AI assistants are gradually but surely entering our daily life.
% To truly assist humans with daily tasks, the ability to understand the world visually and interact with humans naturally is essential. Besides, uncertainties and ambiguities are ubiquitous in the real world. Therefore, AI agents should maintain an active attitude to communicate with humans for a better decision when encountering ambiguity in a target task. 
% To this end, this paper focuses on interactive visual grounding tasks \cite{invig,guesswhat_game} that often occur in real life. For example, as shown in Figure \ref{paradigm}(a), when helping others find an object, it can be difficult to locate the target object due to the ambiguity of the given reference expression. In this case, we people tend to achieve disambiguation first through communication. Correspondingly, we expect AI agents to also be able to disambiguate the reference expression for accurate grounding by holding a dialog with humans naturally.
Robots are increasingly entering our daily lives.
To work with non-expert users, they are required to understand the world visually and interact in languages.
Nevertheless, ambiguities and uncertainties are ubiquitous.
Therefore, robots need to actively seek help from humans to make informed decisions. 
This paper focuses on interactive visual grounding tasks \cite{shridhar2020ingress}, in which robots ask questions to disambiguate and locate targets by interaction, as shown in Fig. \ref{fig:fig_1}.
It poses challenges of 1) how to generate informative questions to gather information based on dialog history and visual observations, and 2) how to ground target objects given much longer dialogues than visual grounding tasks \cite{yu2016modeling, yu2018mattnet}.
Previous approaches usually rely on predefined interactive templates for disambiguation \cite{shridhar2020ingress, zhanglu2021invigorate, yang2022interactive}, which usually requires users to follow rules to interact and are less flexible and robust against open-world scenarios.
% 1. 有信息的问题生成 2. 对长上下文进行grounding目标物体

To address this issue, we present Three-in-One (\tio), a unified model towards end-to-end and robust interactive visual grounding for HRI in the open world.
\tio is trained on a joint of extensive public data, and show superior generality to diversified and challenging open-world scenarios.
To do so,
our model differentiates sub-tasks with different prompts and a shared vocabulary including (object-based) caption generation, (multi-turn) visual question-answering, (multi-turn) visual question generation, visual dialog, and interactive visual grounding.
On this basis, all sub-tasks are unified in a sequence-to-sequence formulation, i.e., outputs are auto-regressively predicted with inputs and previous outputs.
As a result, \tio can respond to different prompts accordingly for different tasks.
% At the inference stage, the model can treat humans as Oracle, thus hold a active dialog with human naturally for disambiguation and accurate grounding. 

\begin{figure*}
  \centering
  % \fbox{\rule[-.5cm]{0cm}{4cm} \rule[-.5cm]{4cm}{0cm}}
  \includegraphics[width=1\textwidth]{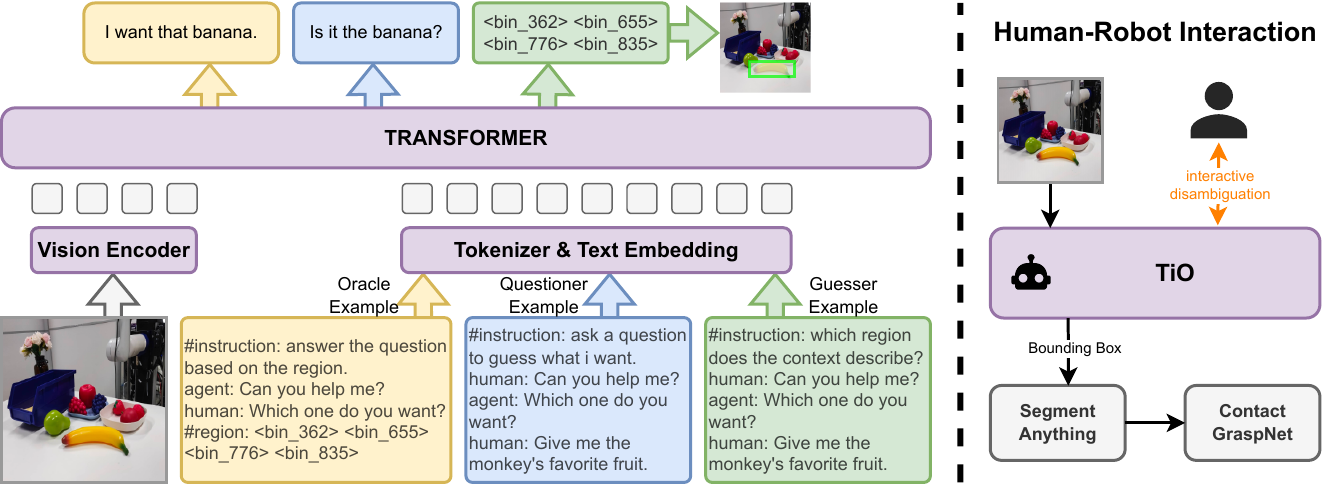} 
  
  \caption{Overview of \tio. Left: \tio network, which is a visual-language interactive disambiguation model that can interact with humans through natural language for disambiguation. It unifies the Questioner, Oracle, and Guesser in a single transformer with different instructions. Right: \tio deployed on interactive manipulation robots. In our interactive manipulation system, \tio provides the target object's bounding box based on the disambiguation by interaction with the human user then converts it into a segmentation map using Segment Anything \cite{kirillov2023segment}. Contact GraspNet \cite{sundermeyer2021contact} finally generates the best grasp based on the projected point clouds.
  }
  \label{fig:overview}
\end{figure*}

We evaluate \tio from three aspects: 1) on standard interactive visual grounding benchmarks, 2) on human-robot interaction with challenging and diversified scenes, and 3) on the real-robot manipulation platform.
For 1), we perform experiments on GuessWhat?!\cite{guesswhat_game} and the more challenging InViG \cite{invig}, where the images are posed with ambiguous instructions and robots are required to ask questions for disambiguation. 
Results show that \tio achieves new state-of-the-art performance of 65.5\% and 78.1\% accuracy, respectively.
Besides, we comprehensively and empirically analyze the superiority of \tio by extensive ablation studies.
For 2), we set up a new challenging benchmark including 150 examples for HRI, to evaluation \tio in the aspect of diversified visual inputs, understanding attributes and behaviors of humans, and open-ended instructions.
We show that \tio outperforms all baseline methods by a clear margin, and demonstrate superior generality to open-world applications.
For 3), we deploy \tio on two real-robot manipulation platforms, one for desktop scenarios following similar settings of \cite{shridhar2020ingress, yang2022interactive}, and another for mobile manipulation tasks.
Results illustrate that \tio achieves robust performance on real-robot applications.

% We summarize our contribution as follows:

% (1) We formulate a simple and effective novel paradigm that unifying Questioner, Oracle and Guesser as a single agent, which is consistent with the behavior of humans asking questions themselves to eliminate ambiguity. 
 
% (2) We propose a unified interactive visual-language disambiguation transformer, which can balance and enhance the abilities of visual dialog and visual grounding, to better handle with interactive visual grounding. %Moreover, it can determine on its own  whether disambiguation finished and when to end the dialog to perform visual grounding, without manually specify.

% (3) Experimental results show the effectiveness of our method, which establishes a new SoTA accuracy on GuessWhat?! and InViG with large margins. In addition, compared with existing methods, experiments also show that our generated dialog when interacting with humans have higher quality. %from the perspective of disambiguation. 

\section{Related Works}

\subsection{Interactive Visual Grounding}

For decision-making in the wild, it is crucial for robots to interact actively when necessary and avoid ambiguity-induced mistakes. 
Classical methods in HRI often rely on predefined corpora to interact actively \cite{ deits2013clarifying,hemachandra2015information,kruijff2006clarification,rosenthal2010effective, tellex2014asking}.
These methods are usually designed for specific tasks and scenarios, limiting their application in real-world interactive scenes.
As deep learning arises, recent advances \cite{johnson2016densecap,liu2017referring, yang2017dense} have shown promise to generate open-ended free-form texts for HRI, which have led to investigations \cite{mo2023towards, shridhar2020ingress, zhanglu2021invigorate, yang2022interactive} of open-ended interaction scenarios in real-world settings.
Nevertheless, the interactions still exhibit limited diversity and naturalness due to a restricted number of templates, making interaction less flexible and even confusing.
To address these issues, recent research has introduced large-scale interactive datasets for more natural HRI \cite{das2017visual, de2017guesswhat, invig}. 
Specifically, GuessWhat?! \cite{de2017guesswhat} focuses on interactive visual disambiguation, where the robot has a visually grounded dialog with humans to identify a target object by iteratively asking judgment questions.
Nevertheless, methods based on GuessWhat?! still suffer from limited interaction patters, setting restrictions for the users to only answer `yes' or `no'.
Recently, InViG \cite{invig} offers more challenging scenarios with 21k open-ended free-form human-human dialogues for interactive visual grounding, which have paved the way for more natural and flexible interactive visual grounding in open-world scenarios.

\subsection{End-to-End and Unified HRI}

With developments in deep learning, visual-language models (VLMs) \cite{lu2019vilbert, su2019vl, lu202012, chen2020uniter, zhang2021vinvl, cho2021unifying, radford2021learning, alayracflamingo, zeng2022xvlm, wang2022ofa, lu2022unifiedio} are more and more prevalent to be the solution of visual-language tasks including Visual Question-Answering (VQA) \cite{antol2015vqa}, Visual Grounding \cite{yu2016modeling}, Image Captioning \cite{chen2015microsoft}, etc.
Nevertheless, these models usually cannot adapt to multi-turn interaction directly.
% Among them, the popular solutions to visual grounding tasks are to first list a large number of bounding boxes proposals, and then select the most relevant one from them. Differently, following \cite{wang2022ofa,lu2022unifiedio}, our method can directly generate the discrete location tokens representing coordinates of the object bounding boxes. 
Recently, large language models have shown promise in language interaction, with impressive few-shot and zero-shot generality to open-ended dialogues \cite{anil2023palm, chowdhery2022palm, brown2020language, du2021glm, touvron2023llama, touvron2023llama2, scao2022bloom, hoffmann2022training}.
Taking advantage of these models, recent works have shown promising results in adapting them to multi-modal inputs \cite{zeng2023matters, zhu2023minigpt4, liu2023visual, dai2023instructblip, ye2023mplug, alayracflamingo}.
Nevertheless, these models are still insufficient to address multi-turn dialogues and follow open-ended instructions in interactive visual grounding, which requires the model to reason from dialog history and visual observations to generate informative questions, and moreover, guess the target objects.

% Recently,taking advantage of  of GPT4 \cite{openai2023gpt4}, has \cite{li2023blip2,huang2023language,dai2023instructblip,zhu2023minigpt4,liu2023visual}. 
% However, due to reliance on LLMs, their output is limited to text format which is not suitable for handling visual grounding tasks. 
% Unlike conventional VLMs and advanced multimodal LLMs, our method can balance both of  human-agent interaction and visual grounding to achieve better visual-language disambiguation.

\section{Preliminaries}

% 这会造成一个drawback。
% The Interactive Visual Dialog task seeks to hold a dialog with the user to disambiguate the expression and identify the intended object. 
Interactive visual grounding means grounding targets visually with ambiguous instructions by iteratively asking questions to gather information.
Formally, a dataset of interactive visual grounding can be denoted by $D = \{I_i, H_i, G_i\}^S_{i=1}$, where $I_i$, $H_i$ and $G_i$ represent the input image, dialog history, and target bounding box of $i-$th sample, respectively.
$S$ is the number of samples in total.
For each sample, $G_i$ is often represented using a 4-d tuple $<x_1, y_1, x_2, y_2>$ indicating the coordinates of upper-left and bottom-right corners. 
The dialog history $H_i=\{Q^{j}_i, O^j_i\}^J_{j=1}$ is a $J$-round conversation between the Questioner and the Oracle.
The Questioner is simultaneously the Guesser, who is to ask questions to guess the target of the Oracle.
The Oracle is the one who knows the intended object and is responsible for response (with $O^j_i$) to the utterances of the Questioner.

% Given the dataset and each input image $I_i$, the goal of the previous paradigm is to train three models as Questioner, Oracle and Guesser to generate $Q^j_i$ based on  $H^{j}_i$, generate  $O^j_i$ based on  $Q^{j-1}_i$, and predict $G_i$ based on complete dialog history $H^J_i$, respectively.  The goal of our paradigm is to train one model with parameter $\theta$ to automatically and properly switch to generate $O^j_i$, $Q^j_i$ or $G^i$ based on the dialog history  $H^j_i$ at each ($j$-th) round conversation. 
% As shown in Figure \ref{paradigm}(b), 
During the inference of a model for interactive visual grounding, previous works \cite{shridhar2020ingress, zhanglu2021invigorate, yang2022interactive} usually rely on a real person to play the role of the Oracle and evaluate the success rate of visual grounding through multi-turn interaction.
Alternatively, an Oracle model can be trained on the same dataset \cite{guesswhat_game} for automatic evaluation.

\section{Method} 

% todo 写明 tio的输入输出很干净, 只有rgb+text

% The previous paradigm adopts independent Questioner, Oracle and Guesser, which leads to Questioner not knowing when Guesser being able to predict correctly and to stop the conversation, resulting in Guesser not controlling the Questioner's dialogue content based on its own understanding. In this section, we introduce our \tio method, which unifies Questioner, Oracle and Guesser into one, as shown in Figure \ref{fig_method}. %is motivated by the human behavior that the quickest way to resolve ambiguity is to personally ask questions based on one's own understanding. 
% We explain more details about architecture, training and inference in Section \ref{3.1}, \ref{3.2} and \ref{3.3}, respectively.

As shown in Fig. \ref{fig:overview}, our \tio is a unified transformer for all visual-language sub-tasks that ensemble interactive visual grounding.
To do so, we 
1) unify training on datasets from image captioning, visual question-answering (VQA), visual grounding (VG), and visual question generation (VQG); 
2) unify prompts and predictions for multiple tasks;
3) unify the encoding and decoding of texts and bounding box coordinates in the tokenizer.
Therefore, during inference, with the corresponding prompt as inputs, \tio can play the role of the Guessor, Oracle, or Questioner, with superior performance compared to baseline methods.
Combined with robot grasping models (e.g. Segment Anything \cite{kirillov2023segment} + Contact GraspNet \cite{sundermeyer2021contact}), \tio can be deployed on the real-robot platform for interactive manipulation tasks robustly with natural language inputs.

\subsection{\tio Network}
\label{3.1}

% \begin{figure}[t]
% \label{fig:overview}
%   \centering
%   % \fbox{\rule[-.5cm]{0cm}{4cm} \rule[-.5cm]{4cm}{0cm}}
%   % \includegraphics[width=0.8\columnwidth]{fig/fig_network.pdf} 
%   \includegraphics[width=0.8\columnwidth]{fig/fig_method.pdf} 
%   \caption{The structure of \tio. }\label{fig_network}
% \end{figure}

% Considering the successful practice of encoder-decoder Transformer \cite{wang2022ofa,raffel2020exploring,cho2021unifying,lewis2019bart} architecture in generative tasks (e.g, dialog generation), we choose it as the backbone architecture of \tio. Model structure is shown in figure \ref{fig_network}.

% \textbf{Vision Encoder.} The image is first scaled to a resolution of 512x512, then a visual encoder is used to convert it into 1024 patch features. To simplify the process of extracting features from input images, ResNet modules are used to encode the images into 1024 patch features. These features are then converted into a hidden space of the same size as the text embedding using a linear layer. 

% \textbf{Tokenizer \& Text Embedding.} The text encoding part involves using BartTokenizer, and then adding 1000 tokens of the bounding box like \cite{wang2022ofa}.

% \textbf{Backbone.} The backbone of the model uses an encoder-decoder Transformer architecture. To represent data from different modalities in a unified space, a strategy of discretizing text, image, and bounding box information is utilized. This involves using tokens in a unified vocabulary. 

% 3. 

% 如何构建我们的模型: 
\paragraph{Backbone} 
Considering the successful practice of encoder-decoder Transformer architecture in generative tasks (e.g., dialog generation, visual grounding) \cite{wang2022ofa,raffel2020exploring,cho2021unifying,lewis2019bart}, we have chosen it as the backbone architecture for \tio, as shown in Fig. \ref{fig:overview}.
% 1. 使用了 transformer+vision_encoder 
% 2. 它的具体构成参数
Our encoder-decoder transformer model within the vision encoder has a parameter capacity of $\sim$930M.
% It uses a hidden size of 1280, a intermediate size of 5120. 
It consists of 24 encoder layers and 12 decoder layers.
The input of the encoder includes both visual and linguistic tokens from vision encoder and text tokenizer.
The decoder is the predictor for all tasks by auto-regressively generating tokens.

\paragraph{Vision Embedding}
% 图片输入处理
The scaled image (target resolution is $512\times512$) is directly used as the input of the vision encoder, then converted into patch features with a 32$\times$32 grid.
Through a learnable linear projection layer, it is then converted into 1024 image embeddings in the hidden space, which are directly fed into the transformers concatenated with text embeddings.

\paragraph{Text Input/Output}
% 文本输入处理
Text tokenization uses BartTokenizer along with adding 1000 location tokens for bounding box prediction.
Formally, each location token is represented by ``<bin\_$i$>'', where $i\in\{0, 1, ..., 999\}$.
% bbox输入处理
The tokenization of object bounding boxes is as follows:
each bounding box $(x_1, y_1, x_2, y_2)$ is firstly normalized and then mapped into the range of $[0, 1000)$.
For example, $(0.0, 0.12, 0.3, 0.4)$ can be converted to "<bin\_0> <bin\_120> <bin\_400> <bin\_300>", and then encoded as location tokens. 
% Bounding box tokenization is usually a part input of \textit{Task:Oracle}.
% bbox输出处理
% bounding box的解码(通常作为Task:Guesser的输出),正好是上述过程的反过程.
% Bounding box detokenization is usually a part output of \textit{Task:Guesser}, which is exactly the reverse of the above process.
% 3. token连接方法
We utilize the identical position encoding as \cite{wang2022ofa}.
Additionally, we combine the image embeddings (red blocks in Fig. \ref{fig:overview}) and text embeddings (blue blocks in Fig. \ref{fig:overview}) together through a simple concatenation. It serves as the input for the transformer's encoder.
The transformer is a sequential-to-sequential architecture, and its decoder predicts one token at a time auto-regressively. 
To perform the generation, we employ the beam-search method with a beam width equal to 5.

\begin{table}
  \caption{ Statistics on the Datasets of Training Tasks}
  \label{tab:datasets}
  \centering
  % \resizebox{\textwidth}{!}{
  \begin{tabular}{llrr}
  
    \toprule
    % \multicolumn{2}{c}{Part}                   \\
    % \cmidrule(r){1-6}
     Source     & Task     & \#Image & \#Sample \\
    \midrule
     SBU Captions\cite{ordonez2011im2text, zhu2023minigpt}  & Caption & 3.4K & 3.4K  \\ 
     LLaVA\cite{liu2023visual}  & VQA & 195.4K & 257.2K  \\ 
     % 67317 + 46965 + 63362 + 17816, 129103 + 46965 + 63362 + 17816
     VisDial\cite{das2017visual}  & Dialog  & 103.4K & 103.4K  \\
     GuessWhat?!\cite{guesswhat_game}     & VG + Dialog & 40.1K & 68.7K \\
     InViG\cite{invig}     &  VG + Dialog & 17.7K & 18.1K \\
     RefCOCO\cite{yu2016modeling} &  VG + Caption & 12.7K & 31.7K    \\
     RefCOCOg\cite{yu2016modeling} &  VG + Caption & 17.4K & 33.1K    \\
    RefCOCO+\cite{yu2016modeling} & VG + Caption & 12.7K & 31.6K \\
    % refcoco 系列 12741 + 17357 + 12740， 31745 + 33068 + 31649
    OpenImages\cite{openimages} & VG &  28.0K & 28.0K\\
   
    %  Task     & Source     & \#Image & \#Sample \\
    % \midrule
    % % Image Caption & SBU, Visual Dialog &&
    
    \bottomrule
  \end{tabular}
  % }
  
\end{table}

%采用多个数据集来训练，分别关注不同的能力，说下数据集的不同，说下组件的不同。

% \begin{figure}
%   \centering
%   % \fbox{\rule[-.5cm]{0cm}{4cm} \rule[-.5cm]{4cm}{0cm}}
%   \includegraphics[width=0.9\textwidth]{training.pdf} 
%   \caption{ A demonstration of the tasks at the training stage, including three task clusters focusing on visual dialog, visual grounding and interactive visual grounding, respectively, to enhance the three abilities (Questioner, Oracle and Guesser) of our Three-in-One model. The datasets involved in model training are selected based on the motivation that we view the abilities of both Questioner and Oracle as the ability to hold a visual dialog, while we view Guesser's abilities as visual grounding.  }\label{training}
% \end{figure}

%说一下整体目标，兼顾两部分能力，三个角色
% A unified framework is designed to 

% During the training phase, in addition to Guesser and Questioner, Oracle is also required. To unify the three into one model, a vital point for the design of model training is the consideration of visual dialog  (for enhancing both Questioner and Oracle abilities), visual grounding (for enhancing Guesser abilities) and interactive visual-language disambiguation (for better coordinating the these abilities). 

\subsection{Training}
\label{sec:training}

% 说一下选择数据集的原因和数量统计
%这里得解释下OpenImages如何做的VG。
\paragraph{Datasets} 
We have shown the datasets involved in training \tio in Table \ref{tab:datasets}.
Apart from GuessWhat?! \cite{guesswhat_game} and InViG \cite{invig} that are inherently suitable for interactive visual grounding, we also collect datasets related to each of the sub-tasks, including Image Captioning (Caption), VQA, Visual Question Generation (VQG), and Visual Grounding (VG).
For abbreviation, we use Dialog to represent the tasks VQA+VQG+Caption+VG.
Concretely, for visual grounding, we collect data from RefCOCO~\cite{yu2016modeling}, and OpenImages~\cite{openimages}.
All target objects are discretized into integer tokens introduced in Section \ref{3.1}.
% Consequently, each bounding box is represented by a 4-dim token sequence $(x_1, y_1,X_2,Y_2)$.
% Each dim is normalized to $[0, 1000]$ by the size of the images.
% Additional processing is applied to these dataset: we format the golden bounding boxes in annotations into the discrete location tokens of our unified vocabulary, as shown in Section \ref{3.1}.  
For image captioning, we collect the Mini-GPT4 Caption \cite{ordonez2011im2text, zhu2023minigpt}, which consists of 3.4K long captions for images.
For dialog, we use LLaVA \cite{liu2023visual} instructions, and Visual Dialog \cite{das2017visual}.
They comprise more than 350K high-quality dialogues and are useful for learning multi-turn language modeling.
Totally, our data includes 955K unique samples based on 135K unique images during the training phase.
% The recent multimodal instruction-tuning data LLaVA is collected for improving the model's generalization ability for responding to various instructions in the real-life human-agent interactive.  Besides, the task-related GuessWhat?! and InVig are essential for the interactive visual-language disambiguation,  because this is the only two datasets that can train all Questioner, Oracle and Guesser capabilities simultaneously.

Notably, one important issue is to train the model to stop conversation after rounds of interaction when information is gathered enough.
To train the model to automatically generate stop signs, we introduce the question ``Can you specify the target object?" during training manually.
The training data is randomly sampled from GuessWhat?! and InViG.
If the sampled dialogue is the last round, the answer will be ``yes'', otherwise, ``No''.
During the inference, we will iteratively and automatically ask the model to answer this question and stops the conversation if the answer is ``Yes''.

\paragraph{Unified Multi-Task Formulation}
Benefiting from the unified transformer architecture introduced in Section \ref{3.1}, it is straightforward to unify all sub-tasks and datasets during training and inference.
Concretely, we differentiate sub-tasks using different prompts as shown in Fig. \ref{fig:overview}.
All tasks are formulated with images and texts as inputs and text-only predictions.
Therefore, our network is trained end-to-end simply using Cross-Entropy loss to maximize the likelihood of the next ground truth token:
\begin{equation}
    loss = \sum_{l=1}^{L} -\log\left[P(w_l | w_{< l})\right]
\end{equation}
where $w_l$ is $l$-th ground truth token for the prediction, $w_{< l}$ is the ground truth token sequence before $w_l$, and $L$ is the length of the ground truth sequence of labels.

\subsection{Interactive Grasping System}
\label{sec:grasp_method}

We deploy \tio with real-robot manipulation tasks.
In our system, \tio is combined with a robot grasping model for interactive visual grounding and grasping.
The system takes raw images and ambiguous text instructions as inputs.
\tio then plays the role of questioner first to interact with humans and gather the necessary information.
When the model stops the conversation actively (see \ref{sec:training} for details), it turns to the role of the Guesser and generates the bounding box of the target object given all dialogue history, which will be fed into the grasping model as the input.
The grasping model consists of two parts: 1) Segment Anything \cite{kirillov2023segment} which is used to get the mask of the target object on 2-D images from the bounding box from \tio, and 2) a Contact GraspNet \cite{sundermeyer2021contact} to detect grasps on the point cloud of the target.
The best grasp with the highest confidence score and no collisions will be selected to execute.

\section{Experiments}

% 为了广泛验证我们提出模型的交互消歧能力, 下面将由浅入深的从三个阶段测评模型. (1) 公开数据集上的性能验证; (2) 人机交互消歧验证; (3) 真实场景下的交互消歧验证.

% In this section, we aim to answer the following questions:
In order to widely evaluate the performance of \tio and other interactive disambiguation methods, this section investigates the following three points:
% we aim to answer the following questions:
\begin{itemize}
% \item What is the performance of \tio on standard benchmarks of interactive visual grounding?
% \item Does \tio generalize well to open-ended interactions with humans?
% \item In what ways can \tio contribute to interactive real-robot manipulation systems?
\item The performance of \tio on standard benchmarks
\item Open-ended interactions with humans
\item The generality to interactive real-robot manipulation systems
\end{itemize}

\subsection{Evaluation on InViG and GuessWhat?!}

\begin{table}[t]
  \caption{Results of Self-play Evaluation on InViG Benchmark}
  \label{end2end2}
  \centering
  \renewcommand{\arraystretch}{1.1}
  %\resizebox{\textwidth}{18mm}{
  \setlength{\tabcolsep}{0.3mm}{
  \scalebox{0.8}{
  \begin{tabular}{l|l|l|c}
  % {
  % >{\centering\arraybackslash}p{30pt}|
  % >{\centering\arraybackslash}X|
  % >{\centering\arraybackslash}X|
  % >{\centering\arraybackslash}X|
  % >{\centering\arraybackslash}p{40pt}
  % }
    \toprule
    % \multicolumn{2}{c}{Part}                   \\
    % \cmidrule(r){1-2}
    % ---------------------------
    % header
    Oracle     & Guesser     & Questioner & SR(\%) \\
    \midrule
    % ---------------------------
    % line 1
    XVLM-Oracle \cite{zeng2022xvlm, invig} & Vilbert-Guesser \cite{tu2021learning} & Vilbert-Questioner \cite{tu2021learning} & 35.3 \\ 
    % ---------------------------
    % line 2
    XVLM-Oracle \cite{zeng2022xvlm, invig}  & XVLM-Guesser \cite{zeng2022xvlm, invig} & XVLM-Questioner \cite{zeng2022xvlm, invig} & 40.1 \\
    \midrule  
    % ---------------------------
    % line 3
    XVLM-Oracle \cite{zeng2022xvlm, invig}  & \multicolumn{2}{c|}{\textbf{\tio} (ours)}  & 46.1 \\ 
    \midrule
    % ---------------------------
    % line 4
    \textbf{\tio} (ours) &Vilbert-Guesser \cite{tu2021learning} & Vilbert-Questioner \cite{tu2021learning} & 51.8 \\
    % ---------------------------
    % line 5
    \textbf{\tio} (ours) &XVLM-Guesser \cite{zeng2022xvlm, invig} & XVLM-Questioner \cite{zeng2022xvlm, invig}  &    61.1 \\
    \midrule
    % ---------------------------
    % line 6
    \multicolumn{3}{c|}{\textbf{\tio} (ours)}  & \textbf{78.1}   \\
    % ---------------------------
    \bottomrule
  \end{tabular}}}
\end{table}

\begin{table}[t]
  \caption{Results of Self-play Evaluation on GuessWhat?! Benchmark}
  \label{end2end1}
  \centering
  \renewcommand{\arraystretch}{1.1}
  % \resizebox{\textwidth}{18mm}{
  \setlength{\tabcolsep}{0.3mm}{
  \scalebox{0.8}{
  \begin{tabular}{l|l|l|c}
    \toprule
    % \multicolumn{2}{c}{Part}                   \\
    % \cmidrule(r){1-2}
    Oracle     & Guesser     & Questioner & SR(\%) \\
    \midrule
    % ---------------------------
    % 其他方法
    Baseline Oracle \cite{guesswhat_game} & Baseline Guesser \cite{strub2017end}  & Baseline Questioner \cite{strub2017end} & 44.6     \\
    Baseline Oracle \cite{guesswhat_game} & Baseline Guesser \cite{strub2017end}  & VDST \cite{pang2019visual} & 45.9     \\
    Baseline Oracle \cite{guesswhat_game} & GDSE-SL \cite{shekhar2018beyond}  & GDSE-SL \cite{shekhar2018beyond} & 47.8     \\
    Baseline Oracle \cite{guesswhat_game} & Guesser(MN) \cite{zhao2018improving}  & TPG \cite{zhao2018improving} & 48.8     \\
    Baseline Oracle \cite{guesswhat_game} & GST \cite{pang2020guessing}  & VDST \cite{pang2019visual} & 50.6     \\
    % Baseline Oracle & Vilbert-Guesser  & VDST & 47.5     \\ 
    % Vilbert-Oracle  & Baseline Guesser & VDST & 47.8 \\
    XVLM-Oracle \cite{zeng2022xvlm, invig}  & XVLM-Guesser \cite{zeng2022xvlm, invig} & XVLM-Questioner \cite{zeng2022xvlm, invig} & 53.0 \\ 
    Vilbert-Oracle \cite{tu2021learning}  & Vilbert-Guesser \cite{tu2021learning} & Vilbert-Questioner \cite{tu2021learning} & 62.8 \\ 
    \midrule
    % \multicolumn{3}{c}{\textbf{TiO} (ours)}  & 57.3   \\
    % ---------------------------
    % 我们的方法
    \multicolumn{3}{c|}{\textbf{\tio} (ours)}  & \textbf{65.5}   \\
    
    % Baseline O
    % \hline
    
    % Axon     & Output terminal & $\sim$10      \\
    % Soma     & Cell body       & up to $10^6$  \\
    \bottomrule
  \end{tabular}}}
\end{table}

% \textit{-GuessWhat/-InViG} represents training with GuessWhat/InViG dataset only. \textit{w/o VD} and \textit{w/o VG.}  represent ``with no visual dialog data'' and ``with no visual grounding data'', respectively. \textit{w/o IVG} represents without interactive visual grounding datasets, i.e.,  GuessWhat?! and InViG.
\begin{table}[t]
  \caption{Ablation Study on InViG Benchmark}
  \label{tab:ablation}

  % \resizebox{\textwidth}{18mm}{
    \begin{tabularx}{\columnwidth}{
    >{\raggedright\arraybackslash}p{70pt}
    >{\centering\arraybackslash}X
    >{\centering\arraybackslash}X}
    \toprule
    % \multicolumn{2}{c}{Part}                   \\
    % \cmidrule(r){1-2}
     Model     & Self-play & Guesser\\
     \midrule
     \textbf{\tio} & 78.1\% & 77.1\%      \\
     \textbf{\tio}-Small & 71.5\% & 74.9\%      \\  
    \midrule
     % \textit{w/ ft.} & X & X      \\ \hline
    \textit{Only GuessWhat?!} & 49.1\% & 56.3\%    \\ 
    \textit{Only InViG} & 56.9\% & 71.2\%    \\
     
    % \textit{w/o ext.} & 56.9 & 71.2  \\ 
     
    \textit{w/o Dialog} & 68.9\% & 75.5\%  \\ 
    \textit{w/o VG} & 65.8\% & 73.7\%  \\  %\hline
    % \ \ \textit{w/o Dialog+VG} & 69.0\% & 73.5\%  \\  \hline
    \textit{w/o IVG} & < 5\% & < 5\%  \\ 
    \bottomrule
    \end{tabularx}
\end{table}

\begin{table}[t]
  \caption{Ablation Study on GuessWhat?! Benchmark}
  \label{tab:ablation_guesswhat}

  % \resizebox{\textwidth}{18mm}{
    \begin{tabularx}{\columnwidth}{
    >{\raggedright\arraybackslash}p{70pt}
    >{\centering\arraybackslash}X
    >{\centering\arraybackslash}X
    >{\centering\arraybackslash}X}
    \toprule
     Model     & Self-play & Guesser & Oracle\\
     \midrule
     \textbf{\tio} & 65.5\% & 74.6\%  & 88.6\%     \\
     \textbf{\tio}-Small & 61.3\% & 72.1\%  & 86.3\%     \\ 
    \midrule
    \textit{Only GuessWhat?!} & 61.8\% & 71.8\% & 86.4\%     \\ 
    \textit{Only InViG} & 37.2\% & 46.9\% & 39.1\%     \\ 
    \textit{w/o Dialog} & 61.3\% & 72.6\% & 86.1\%     \\ 
    \textit{w/o VG} & 56.2\% & 71.1\% & 84.7\%   \\ %\hline
    \textit{w/o IVG} & < 5\% & < 5\%  & < 5\%  \\    
    \bottomrule
    \end{tabularx}
\end{table}

\paragraph{Experiment Setup} 
Firstly, we evaluate \tio and baselines based on \textit{Self-play Success Rate} on two benchmarks: InViG and GuessWhat?!.
InViG \cite{invig} is an interactive visual grounding dataset that contains 21k open-ended free-form human-human dialogues and 17.7k images with several ambiguous objects in each image. 
% It is not limited to the question-answer form.
GuessWhat?! \cite{guesswhat_game} is a simpler dataset that only contains close-ended questions with answers limited to "yes", "no", and "n/a". 
It consists of 155k dialogues on 66k images, totaling 821k question-answer pairs.
In the self-play evaluation, the input contains an image of the scene and an ambiguous language instruction.
The interactive model will simultaneously play the role of Questioner, Oracle, and Guesser, to iteratively ask informative questions, answer the questions by itself, and finally stop the conversation and guess the target from the dialogue history.
A prediction is considered to be correct if it has an Interaction of Union (IoU) larger than 0.5 with the ground truth.
Self-play offers chances for large-scale evaluation of thousands of interactive examples without the participation of humans, while being challenging and comprehensive to simultaneously evaluate the joint performance of all sub-tasks.
% We evaluated the performance of \tio using a \textit{self-play} setting. In the disambiguation task setting, the Questioner and Guesser are the roles of robots, while the Oracle plays the role of a human. Self-play refers to the interaction between the Questioner\&Guesser model and the Oracle model without human participation. 
% Unlike previous methods that trained separate models for the Questioner, Guesser, and Oracle, we merged these roles into a single model and distinguished them by prompt.

\paragraph{Compared to Baselines} 
The performance comparison on the InViG benchmark is shown in Table \ref{end2end2}. When XVLM-Oracle is as an Oracle, \tio surpassed Vilbert-Guesser/Questioner by 10.8\% in disambiguation success rate and surpassed XVLM-Guesser/Questioner by 6.0\%. When \tio is as an Oracle, \tio surpassed Vilbert-Guesser/Questioner by 26.3\% in disambiguation success rate and surpassed XVLM-Guesser/Questioner by 17.0\%. The self-play disambiguation success rates of \tio (as an Oracle) relative to XVLM-Oracle increased by 16.5\%, 21.0\%, and 32.0\% for Vilbert, XVLM, and \tio, respectively.
The performance comparison on the GuessWhat?! benchmark is shown in Table \ref{end2end1}. \tio's self-play success rate surpasses the best baseline by 2.7\%. Note, the answer is limited to "yes", "no", and "n/a" on GuessWhat?!, and there is still a certain gap with the dialogue in the real scene. 
These results demonstrate that the joint training strategy of \tio can comprehensively improve the visual language alignment ability and the context understanding ability of the Oracle, Questioner, and Guesser within the scope of these two benchmarks. 

\begin{figure*}[t]
  \centering
  % \fbox{\rule[-.5cm]{0cm}{4cm} \rule[-.5cm]{4cm}{0cm}}
  
  \includegraphics[width=0.92\textwidth]{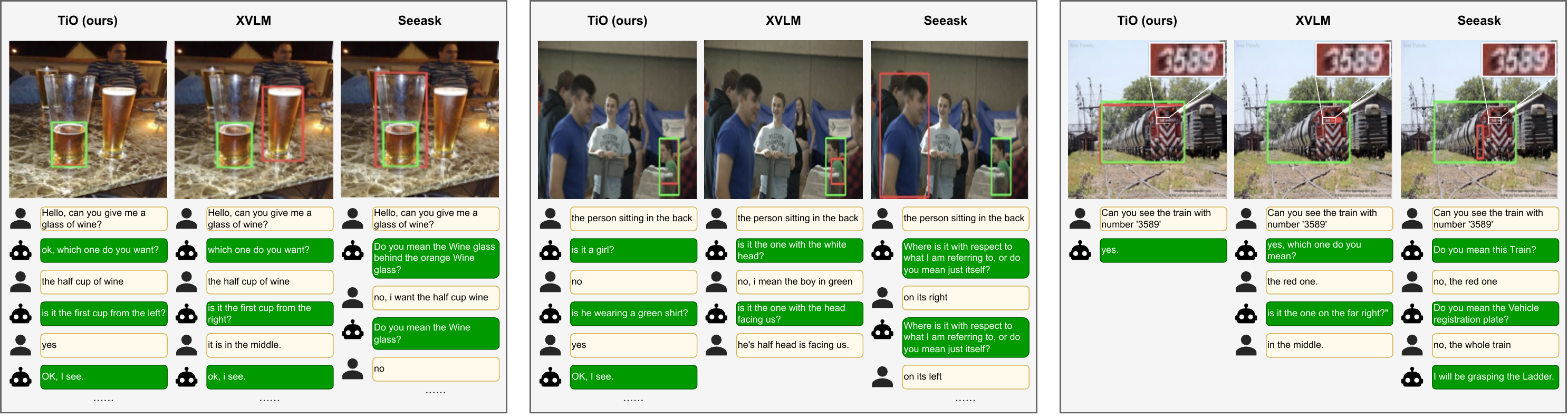} 
  
  \caption{Qualitative results of different interactive visual grounding methods on our 3 HRI evaluation sets. Left: \textit{scene understanding}. Middle: \textit{human understanding}. Right: \textit{language understanding}. The green box denotes the target object by the human user, and the red box denotes the prediction after interaction.
  }\label{exp2_1_vis}
\end{figure*}

\begin{figure}[t]
  \centering
  % \fbox{\rule[-.5cm]{0cm}{4cm} \rule[-.5cm]{4cm}{0cm}}
  
  \includegraphics[width=0.9\columnwidth]{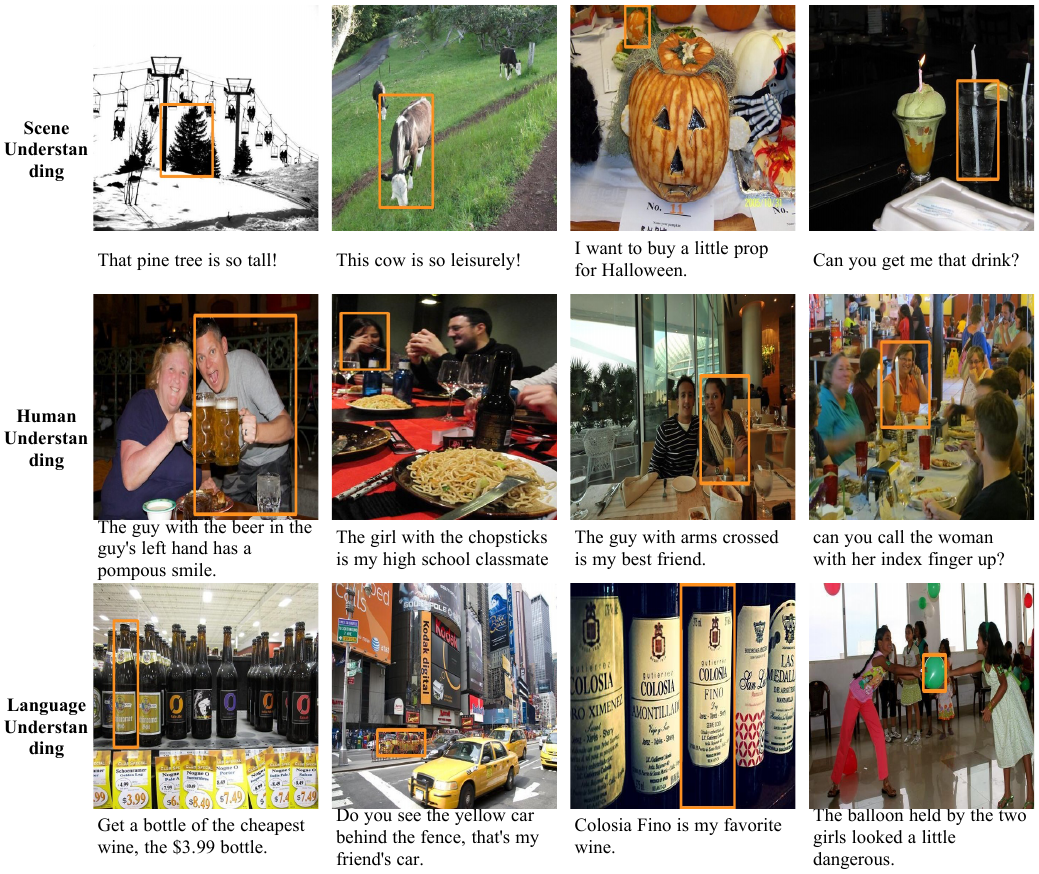} 
  
  \caption{Examples of our evaluation benchmark for HRI experiments. Top row: \textit{Scene Understanding}. Middle row: \textit{Human Understanding}. Bottom row: \textit{Language Understanding}.}\label{evaluation_set}
\end{figure}

\paragraph{Ablation Studies} We also conduct ablation studies on GuessWhat?! and InViG to figure out the contributions of unified training.
To be specific, we first implement \tio with a smaller capacity (\tio-Small) with $\sim$470M parameters, about half of the full version.
As shown in Table \ref{tab:ablation} and Table \ref{tab:ablation_guesswhat}, we can conclude that model capacity matters.
Larger models consistently improve the performance of all sub-tasks.
We also ablate different data sources with \tio-Small to validate their effects on the performance.
Concretely, \textit{Only GuessWhat?!} and \textit{Only InViG} mean training on a single interactive dataset.
\textit{w/o Dialog} indicates without VisDial and LLaVA.
\textit{w/o VG} means without RefCOCO and OpenImages.
\textit{w/o IVG} means without both GuessWhat?! and InViG.
% todo 缺乏数值描述
We can conclude that:
1) More sub-tasks improve performance significantly;
2) The model trained solely on GuessWhat?! or InViG performs relatively well in its own domain, but generalizes worse than the full version.
3) Interactive visual grounding data itself is indispensable. Models cannot achieve reasonable performance without such data.
4) Dialog and VG data can further improve performance when combined with the interactive visual grounding data.

\begin{figure}[t]
  \centering
  % \fbox{\rule[-.5cm]{0cm}{4cm} \rule[-.5cm]{4cm}{0cm}}
  
  \includegraphics[width=1\columnwidth]{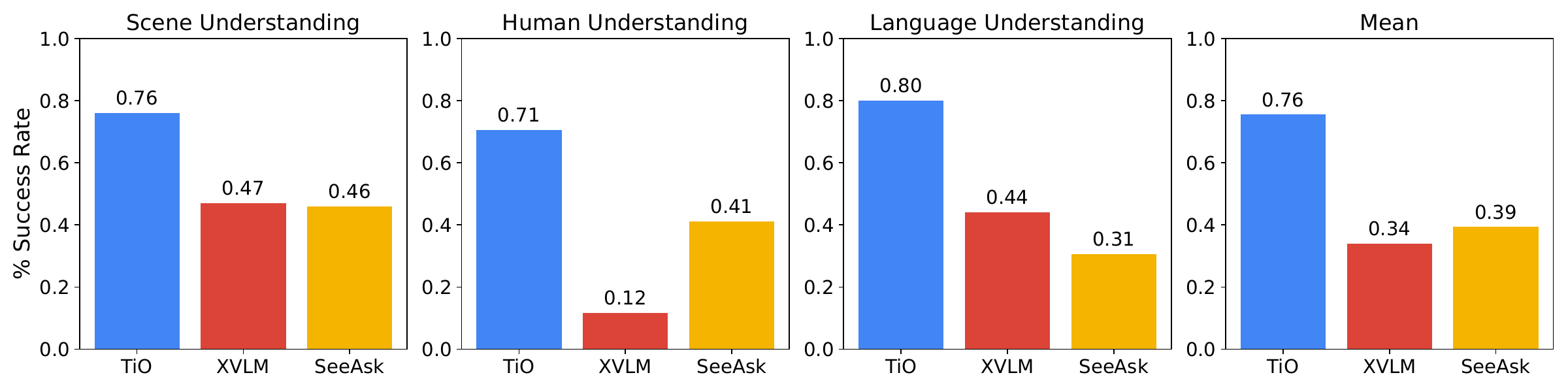} 
  
  \caption{Interactive visual grounding success rate of HRI on 3 evaluation sets. Our approach achieves the highest performance on the more challenging interactive scenarios. }\label{exp2_1}
\end{figure}

\begin{figure}[t]
  \centering
  % \fbox{\rule[-.5cm]{0cm}{4cm} \rule[-.5cm]{4cm}{0cm}}
  \includegraphics[width=0.9\columnwidth]{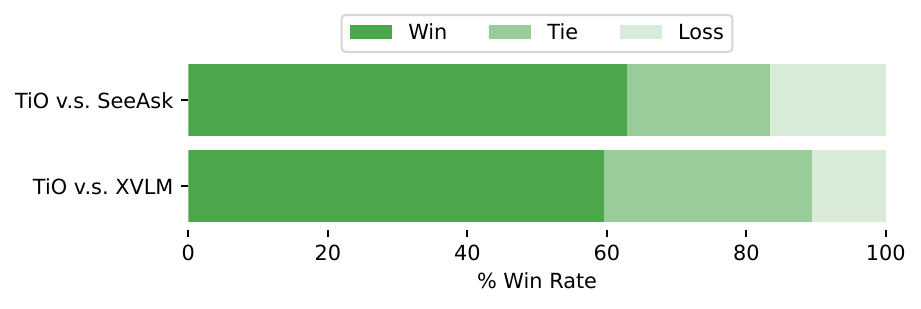} 
  \caption{Human scoring of \tio against baselines. }\label{exp2_win_rate}
\end{figure}

\subsection{Evaluation on Human-Robot Interaction}

% 为了更好的评估消歧模型的上下文理解能力和视觉语言对齐能力,
\paragraph{Experimental Setup}
% For 2), we sample a challenging test set from the test set of the InViG dataset, \textit{openimages}, and \textit{objects365}.
% Based on the selected images, we then recruit 10 volunteers to interact with \tio.
% To make the evaluation comprehensive, we evaluate the performance of \tio on understanding diversified visual inputs (\textit{Visual Understanding}), human attributes \& behaviors (\textit{Human Understanding}), and language expressions (\textit{Language Understanding}), which are usually required in open-ended HRI applications.
% For each one of them, we select 50 images and re-label the instructions before the interaction experiments.
% Therefore, we have 150 samples in total in this phase.
% Some examples are shown in Fig. \ref{evaluation_set}.

% 为了更加全面的验证TiO的消歧能力, 我们提出了一个极具挑战消歧验证集合. 这个验证集合包含150张图片, 
In order to more comprehensively evaluate the disambiguation ability of \tio, we propose a challenging disambiguation evaluation set, which contains 150 images from the test set of the InViG dataset, \textit{OpenImages} \cite{openimages}, and \textit{objects365} \cite{shao2019objects365}, 50 of which are sampled from the images containing \textit{human}-related categories. Then, we divide it into 3 parts aim to evaluate the performance of models on understanding diversified visual concepts (\textit{Scene Understanding}), human attributes \& behaviors (\textit{Human Understanding}), and language expressions (\textit{Language Understanding}), which are usually required in open-ended HRI applications. For each one of them, we select 50 images and re-label the instructions. Based on the evaluation set, we then recruit 10 volunteers to interact with \tio and baselines.

\paragraph{Compared to Baselines} The success rate of interactive visual grounding with the volunteers on the 150 samples set for HRI is shown in Fig. \ref{exp2_1}.
% 整体上看,我们的方法在三个验证集上均优于XVLM和SeeAsk. TiO使用了更多视觉语言数据,因此具有强大的泛化性. 在三个集合上分别取得了76%,70.6%,80%的成功率. XVLM在"human understanding" 集合上的性能明显偏低,是因为invig训练集中"人"相关的数据偏少导致的(<5%). SeeAsk作为一种rule-base的方法在三个验证集上具有相对稳定的性能, 但它在"语言理解"的验证集上的性能相对偏低. 它使用了part-parser和clip来进行结构化的语言理解,因此在面对复杂表达时就显得力不从心了.
Our method outperforms XVLM \cite{zeng2021multi} and SeeAsk \cite{mo2023towards} on all 3 evaluation subsets.
We achieved success rates of 76.0\%, 70.6\%, and 80.0\% on the sets of Scene Understanding, Human Understanding, and Langauge Understanding, respectively.
By contrast, XVLM performs significantly worse on the \textit{human understanding} set due to the scarcity of human-related data in InViG dataset (<1\%).
SeeAsk, being a rule-based method, demonstrates relatively stable performance on the 3 evaluation subsets.
However, its performance is relatively low on the Language Understanding subset, revealing the drawbacks of rule-based methods on understanding diversified and open-ended language inputs. A human scoring based on the evaluation of the disambiguation dialog is shown in Fig. \ref {exp2_win_rate}. It can be found that, the dialog between humans and \tio is more preferred than 2 baselines in most cases.
% SeeAsk employs part-parser and clip for structured language understanding\cite{mo2023towards}, making challenges in complex expressions.

% 人类根据消歧对话历史的有效性的评分结果 is shown in Fig. \ref{exp2_win_rate}. 可以发现, 在大部分样本中, 人类和TiO的对话都优于和baselines的对话. 

\paragraph{Qualitative Demonstrations} As shown in Fig. \ref{exp2_1_vis}, we can see that \tio exhibits strong generalization to diversified inputs and open-world settings.
It can understand fine-grained linguistic concepts, including ``half cup of wine'' against ``cup'' in the first example, and ``with the number 3589'' in the third example.
Besides, \tio can generate informative and natural questions to exclude disturbance efficiently.
By contrast, SeeAsk can only follow question templates without diversified language outputs.

% todo 差一个评估阶段对比性能

% ---------------------------------------------------
% □ 实验三: Real-World Evaluation
% ---------------------------------------------------
% 图表: 抓取成功率一个
\subsection{Real-World Evaluation}

\begin{figure}[t]
  \centering
  % \fbox{\rule[-.5cm]{0cm}{4cm} \rule[-.5cm]{4cm}{0cm}}
  \includegraphics[width=1\columnwidth]{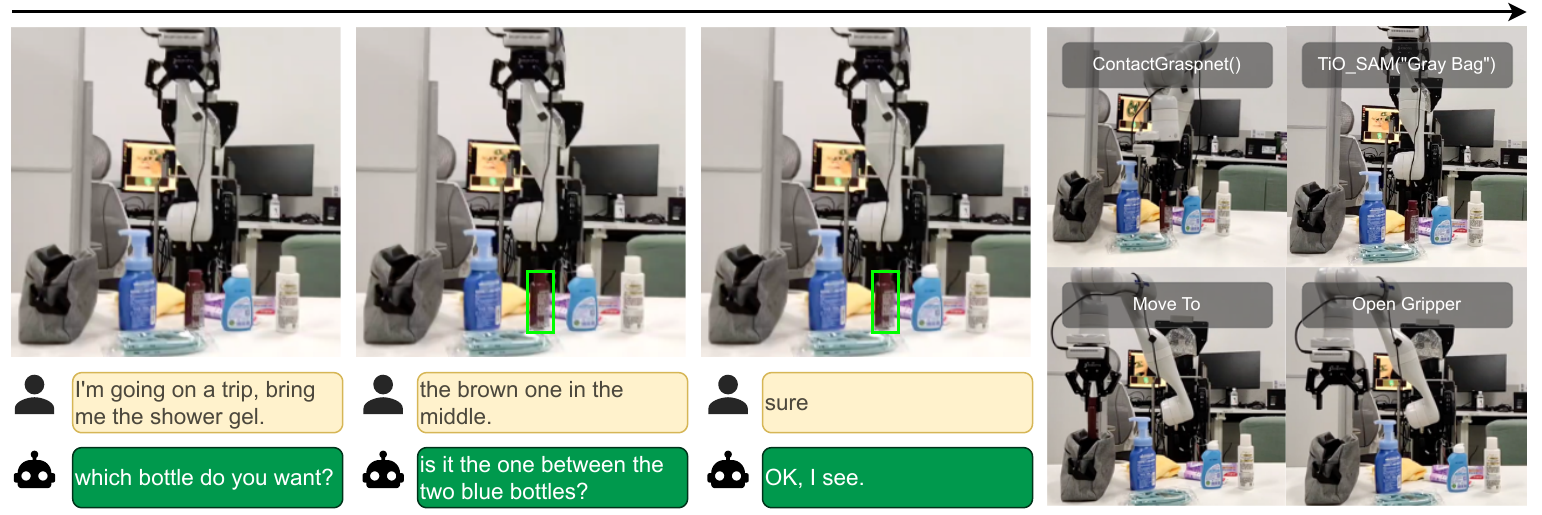} 
  \caption{Showcase of the interactive visual grounding in robotic manipulation tasks on the desktop.}\label{exp_real_robot}
\end{figure}

\begin{figure}[t]
  \centering
  % \fbox{\rule[-.5cm]{0cm}{4cm} \rule[-.5cm]{4cm}{0cm}}
  \includegraphics[width=1\columnwidth]{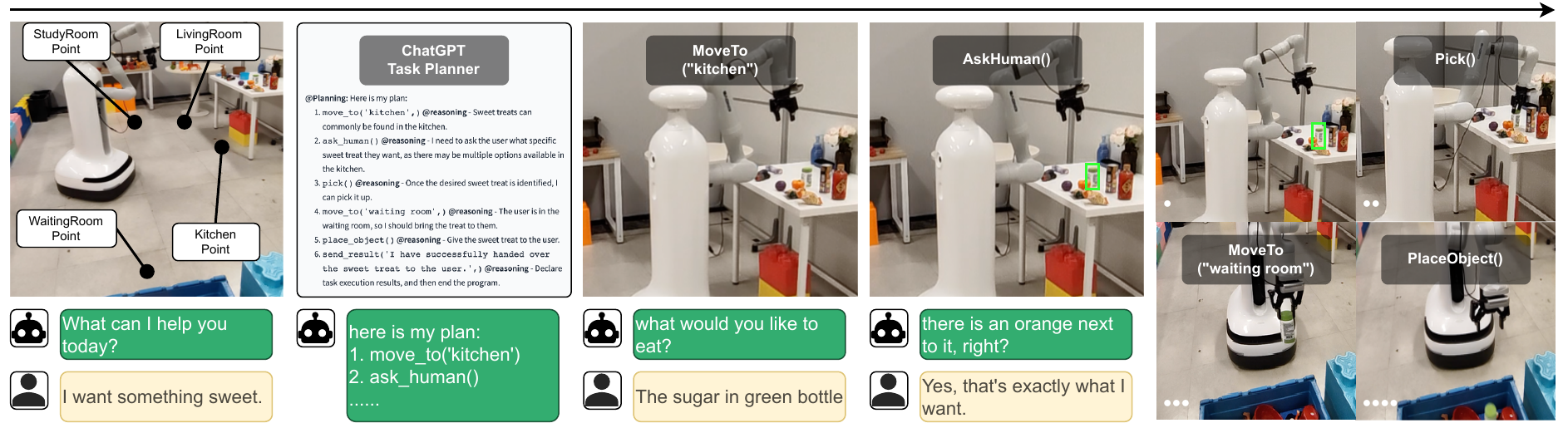} 
  \caption{Showcase of the interactive visual grounding in robotic manipulation tasks on the mobile platform. }\label{exp_real_mobile}
  \label{fig:real_robot_2}
\end{figure}

% \section{Limitation and Future Works}
% balabalabala...
\begin{table}[t]
  \caption{Disambiguation and Grasping success rate in desktop scenes.}
  \label{table_real_grasp}
  
  % \resizebox{\textwidth}{18mm}{
    \begin{tabularx}{\columnwidth}{>{\raggedright\arraybackslash}p{40pt}|>{\centering\arraybackslash}X|>{\centering\arraybackslash}X}
    \toprule
    Method & Grounding Success Rate & Grasp Success Rate \\  
    \midrule
    XVLM & 40\% (6/15) & 40\% (6/15) \\  
    SeeAsk & 46\% (7/15) & 46\% (7/15) \\  
    \tio & \textbf{86\% (13/15)} & \textbf{73\% (11/15)} \\  
    \bottomrule
    \end{tabularx}
    \label{tab:real_robot_desktop}
\end{table}

% \begin{figure}[t]  
%   \centering  
%   \includegraphics[width=0.6\columnwidth]{fig/real_robot.jpg}  
%   \caption{Robot platform used for interactive robot manipulation.}  
%   \label{fig:robot_platform}  
% \end{figure} 

% these findings demonstrate the versatility and efficacy of \tio when deployed on indoor mobile robots. The successful integration of \tio into mobile robotic platforms opens up possibilities for enhanced HRI in a wide range of real-world scenarios.

% 为了验证在真实人机交互场景中我们方法的消歧性能, 我们将相关算法部署在桌面协作机械臂和室内移动机器人上. 我们先汇报XVLM, SeeAsk, TiO放在桌面场景下的消歧成功率和抓取成功率, 然后展示TiO部署在室内移动机器人上的效果.

\paragraph{Experimental Setup}
% KINOVA GEN3
% Kinova Gen3 is an advanced robotic arm. It adopts advanced sensing technology and intelligent control system to realize precise and flexible movement execution. The robotic arm has a wide range of applications, including industrial automation, medical assistance, scientific research and experimentation. Users can control it through a variety of programming languages and development environments for personalized operation and task planning. Whether you are a beginner or a professional, Kinova Gen3 offers strong support and innovative solutions. Whether you are a beginner or a professional, Kinova Gen3 provides strong support and innovative solutions.
In this phase, we have deployed \tio on two real-robot platforms to evaluate the performance of interactive robotic manipulation.
% The mobile robot for manipulation used in our experiments is shown in the first image of \ref{fig:real_robot}.
We apply a Kinova arm for manipulation and a RealSense camera for RGB and point cloud observation.
The mobile platform is developed by our own team.

\paragraph{Disambiguation on Desktop Robot}
% In this section, we present the disambiguation success rate and grasp success rate achieved by three different methods: TiO, XVLM, and SeeAsk, in the context of a desktop scene.
Before the interaction, the robot observes the scene from above the desktop.
After receiving the instruction, it interacts with the user to gather information, after which it invokes the grasping model (Section \ref{sec:grasp_method}) for grasp execution.
% 抓取实验
% 我们在15个场景中测试了三种方法的消歧成功率和抓取成功率, 
% 实验结果如表所示. 从表中可以看出, 
% todo 配图消歧成功率和抓取成功率
To evaluate the effectiveness of the disambiguation process, we measure the success rate of each method in correctly understanding the user’s intent (Table \ref{tab:real_robot_desktop}). 
Our method, TiO, outperforms both XVLM and SeeAsk by a large margin, achieving the highest interactive grounding success rate of 86\%. 
This result demonstrates the robustness and accuracy of \tio in understanding and disambiguating the user’s ambiguous request.
% The grasping success rate is also a crucial metric for evaluating the ability of a robotic system to accurately and reliably grasp objects. 
% Therefore, as shown in Table \ref{tab:real_robot_desktop}, 
We also compare the grasp success rates of TiO, XVLM, and SeeAsk in the desktop scene. 
Remarkably, \tio exhibited the highest grasp success rate among the three methods. 
It highlights the efficacy of our system in enabling the robot to successfully interact with objects physically in the desktop environment.

\paragraph{Disambiguation on Mobile Robot}
% Effect of \tio Deployment on Indoor Mobile Robots
By integrating \tio into mobile robotic platforms, we showcase its performance and adaptability in more complex tasks. 
% In this phase, we evaluate the performance of \tio on various mobile robotic platforms operating in indoor settings. 
As shown in Fig. \ref{fig:real_robot_2}, In this phase, \tio is combined with a task planner for navigation in a house.
In detail, after receiving the instruction from the human user, the task planner (ChatGPT in our case) will plan for a macro-action sequence given the instruction and a list of macro-action candidates (e.g. move\_to, ask\_human, grasp).
The robot will follow the plan to navigate to the first place.
The \tio is responsible for the multi-modal interaction and disambiguation with humans.
Concretely, it first observes the scene and detects whether possible candidates exist.
If the answer is ``Yes'', it continues to interact with humans.
Otherwise, the task planner will be invoked again to re-plan for the next place.
The existence detection of candidates is trained with the question ``Is there a [OBJ]?'', and the training data is generated directly from OpenImages.
Our results indicate that \tio can successfully facilitate the disambiguation process, enabling the robot to accurately interpret human intentions and execute the corresponding actions in dynamic scenarios. 
% Furthermore, \tio showcased a remarkable grasp success rate, further validating its ability to interact with objects in the mobile robot’s surroundings.

% 真机实验证明了我们的方法在真实世界具有

% \paragraph{Overall}
% Overall, these findings demonstrate the versatility and efficacy of \tio when deployed on indoor mobile robots. The successful integration of \tio into mobile robotic platforms opens up possibilities for enhanced HRI in a wide range of real-world scenarios.

\section{Conclusion}

This paper proposes \tio that unifies three agents Questioner, Oracle, and Guesser into one single transformer, 
%namely Three-in-One (TiO), 
for interactive visual grounding.
% It is easier to deploy and more in line with human behavior than the three-agent solution. On this basis, we propose an interactive visual-language disambiguation Transformer that can hold a natural and informative dialog with human users to disambiguate their expression and identify the target object. 
% Benefiting from the unified formulation of visual-language sub-tasks, \tio can be trained on a joint of extensive public datasets and achieves robust performance in open-ended interactive visual grounding.
Compared with baselines including XVLM \cite{zeng2021multi} and SeeAsk \cite{mo2023towards}, \tio sets the new state-of-the-art both on standard benchmarks of GuessWhat?! and InViG, and moreover, the interaction with humans, outperforming baselines by a large margin.
We also deploy \tio on real-robot platforms and validate its effectiveness in real-world interactive robotic manipulation tasks.
Currently, restricted by the training data, \tio can only generate simple sentences for interaction.
In the future, it is promising to improve the text generation ability by using text datasets.
Besides, it is also important for \tio to adapt to more expressions of robotic manipulation (e.g. the grounding of actions), which is important for interactive manipulation systems.

% We conduct extensive experiments on GuessWhat?! and challenging InViG, which demonstrates the excellence of the proposed paradigm and Transformer. \tio can also generalize to real-world products, e.g., household robot, as human evaluation proves that it is more user-friendly during the interaction with humans. 


%%%%%%%%%%%%%%%%%%%%%%%%%%%%%%%%%%%%%%%%%%%%%%%%%%%%%%%%%%%%%%%%%%%%%%%%%%%%%%%%

%%%%%%%%%%%%%%%%%%%%%%%%%%%%%%%%%%%%%%%%%%%%%%%%%%%%%%%%%%%%%%%%%%%%%%%%%%%%%%%%



% \section*{ACKNOWLEDGMENT}
\section*{ACKNOWLEDGMENT}

This work was supported in part by National Key R\&D Program of China under grant No. 2021ZD0112700, the NSFC under grant No. 62125305, No. U23A20339, No. 62088102, No. 62203348, and No. 61973246.

We would like to acknowledge Hang Li at ByteDance for his generous assistance and insightful comments in technical discussions. 
Additionally, we extend our appreciation to the volunteers who have been involved in the human-robot interaction experiments.

% The preferred spelling of the word ÒacknowledgmentÓ in America is without an ÒeÓ after the ÒgÓ. Avoid the stilted expression, ÒOne of us (R. B. G.) thanks . . .Ó  Instead, try ÒR. B. G. thanksÓ. Put sponsor acknowledgments in the unnumbered footnote on the first page.

%%%%%%%%%%%%%%%%%%%%%%%%%%%%%%%%%%%%%%%%%%%%%%%%%%%%%%%%%%%%%%%%%%%%%%%%%%%%%%%%

{
    % https://www.overleaf.com/learn/latex/Bibtex_bibliography_styles
    \small
    \bibliographystyle{unsrt}
    \bibliography{bib/my-bib.bib}
}

% \begin{thebibliography}{99}

% \end{thebibliography}

\end{document}